\documentclass[10pt,twocolumn,letterpaper]{article}

\usepackage{cvpr}              

\usepackage{graphicx}
\usepackage{amsmath}
\usepackage{amssymb}
\usepackage{booktabs}
\usepackage{stfloats}
\usepackage{multirow}
\usepackage{multicol}
\usepackage{makecell}
\usepackage{diagbox}
\usepackage[accsupp]{axessibility}

\usepackage[pagebackref,breaklinks,colorlinks]{hyperref}

\usepackage[capitalize]{cleveref}
\crefname{section}{Sec.}{Secs.}
\Crefname{section}{Section}{Sections}
\Crefname{table}{Table}{Tables}
\crefname{table}{Tab.}{Tabs.}

\def\cvprPaperID{71} 
\def\confName{CVPR}
\def\confYear{2023}

\begin{document}

\title{BokehOrNot: Transforming Bokeh Effect with Image Transformer \\ and Lens Metadata Embedding}

\author{Zhihao Yang \, Wenyi Lian \, Siyuan Lai \\
Uppsala University, Sweden\\
{\tt\small \{zhihao.yang.8094,wenyi.lian.7322,siyuan.lai.3205\}@student.uu.se}}


\maketitle

\begin{abstract}
   Bokeh effect is an optical phenomenon that offers a pleasant visual experience, typically generated by high-end cameras with wide aperture lenses. The task of bokeh effect transformation aims to produce a desired effect in one set of lenses and apertures based on another combination. Current models are limited in their ability to render a specific set of bokeh effects, primarily transformations from sharp to blur. In this paper, we propose a novel universal method for embedding lens metadata into the model and introducing a loss calculation method using alpha masks from the newly released Bokeh Effect Transformation Dataset (BETD)~\cite{conde2023ntire_bokeh}. Based on the above techniques, we propose the BokehOrNot model, which is capable of producing both blur-to-sharp and sharp-to-blur bokeh effect with various combinations of lenses and aperture sizes. Our proposed model outperforms current leading bokeh rendering and image restoration models and renders visually natural bokeh effects. Our code is available at: \url{https://github.com/indicator0/bokehornot}.
\end{abstract}

\section{Introduction}
\label{sec:intro}
\begin{figure}
  \centering
  \begin{subfigure}{0.32\linewidth}
    \includegraphics[width=1.1in]{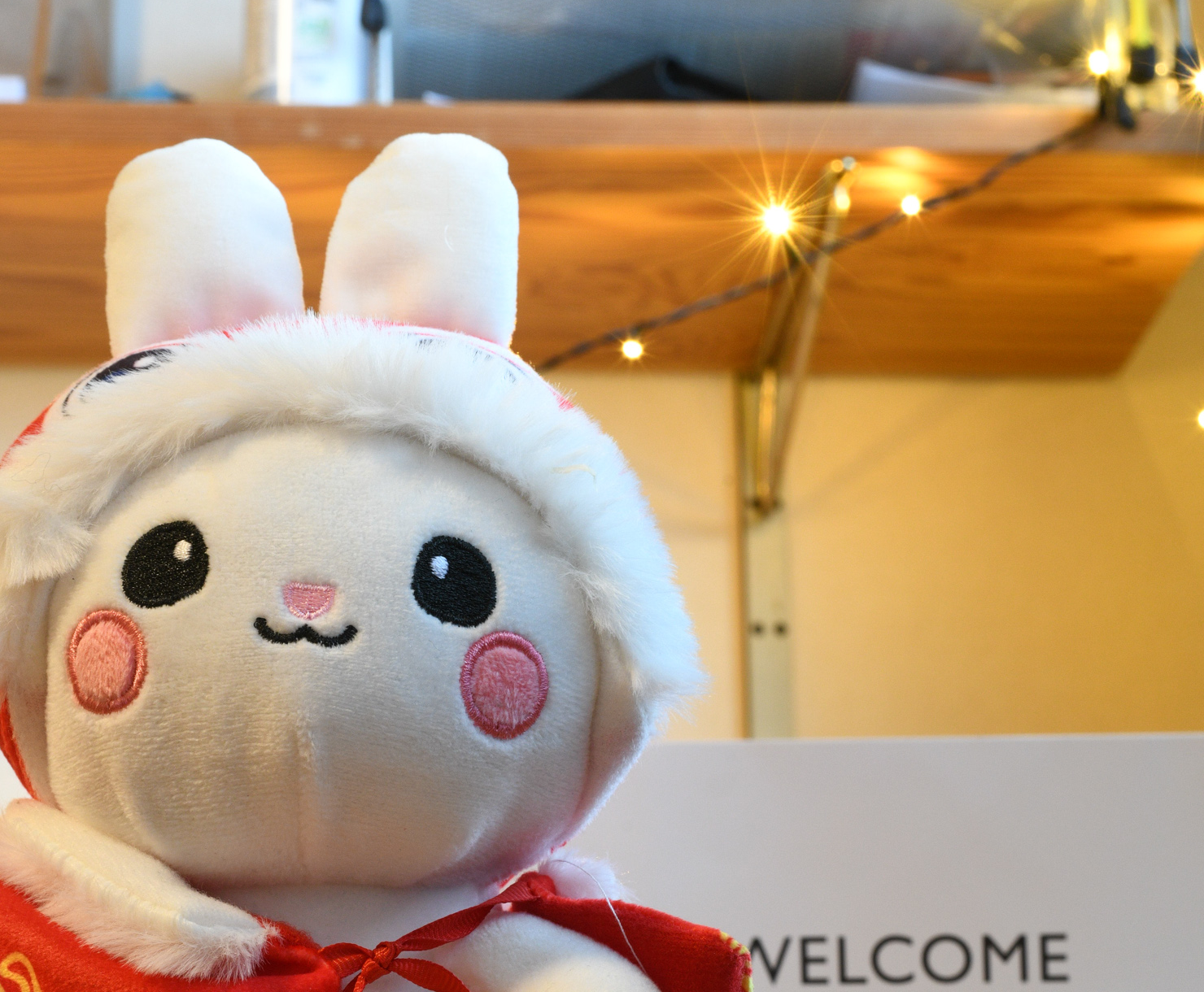}
    \caption{}
    \label{fig:intro-a}
  \end{subfigure}
  \hfill
  \begin{subfigure}{0.32\linewidth}
    \includegraphics[width=1.1in]{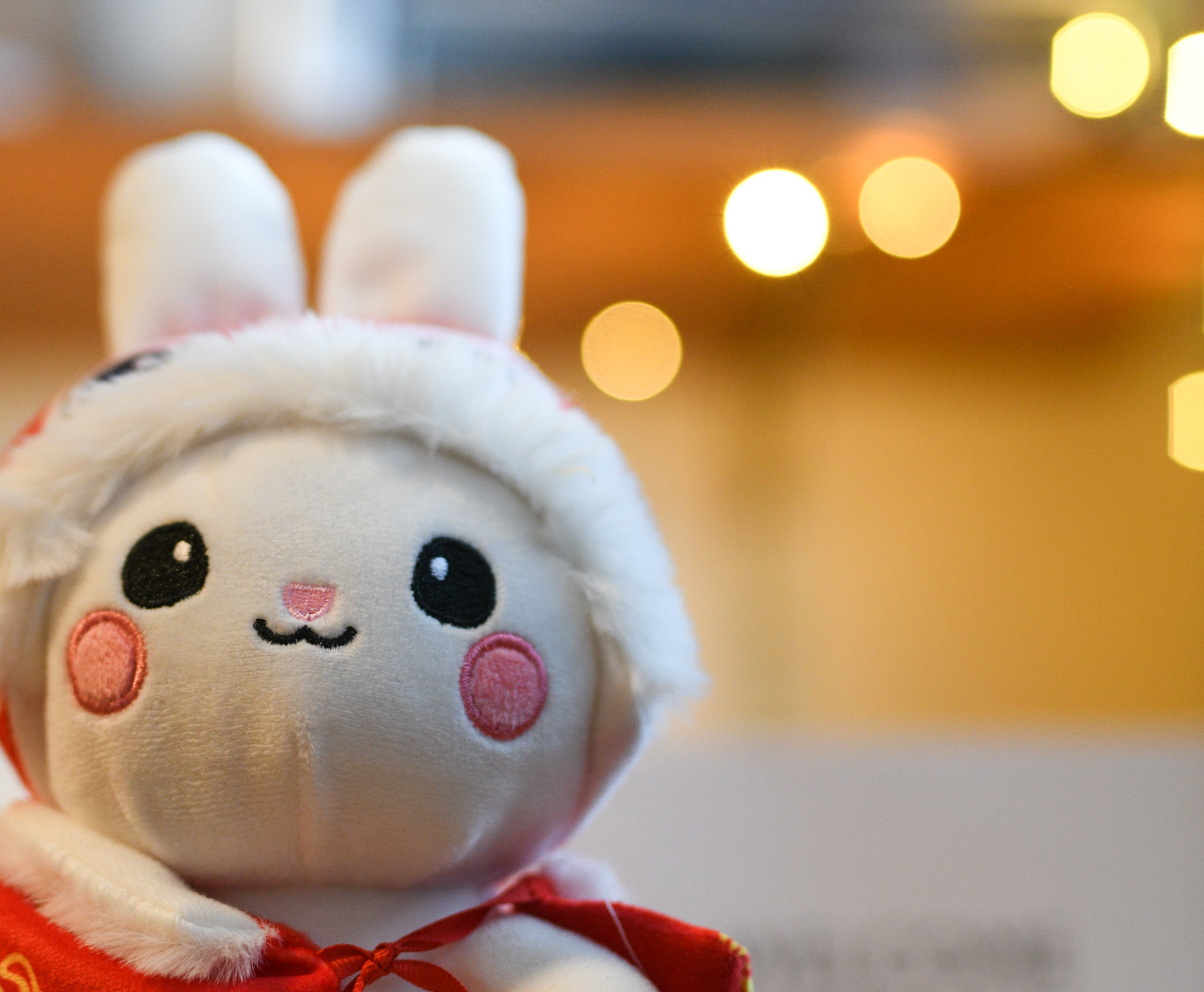}
    \caption{}
    \label{fig:intro-b}
  \end{subfigure}
  \hfill
  \begin{subfigure}{0.32\linewidth}
    \includegraphics[width=1.1in]{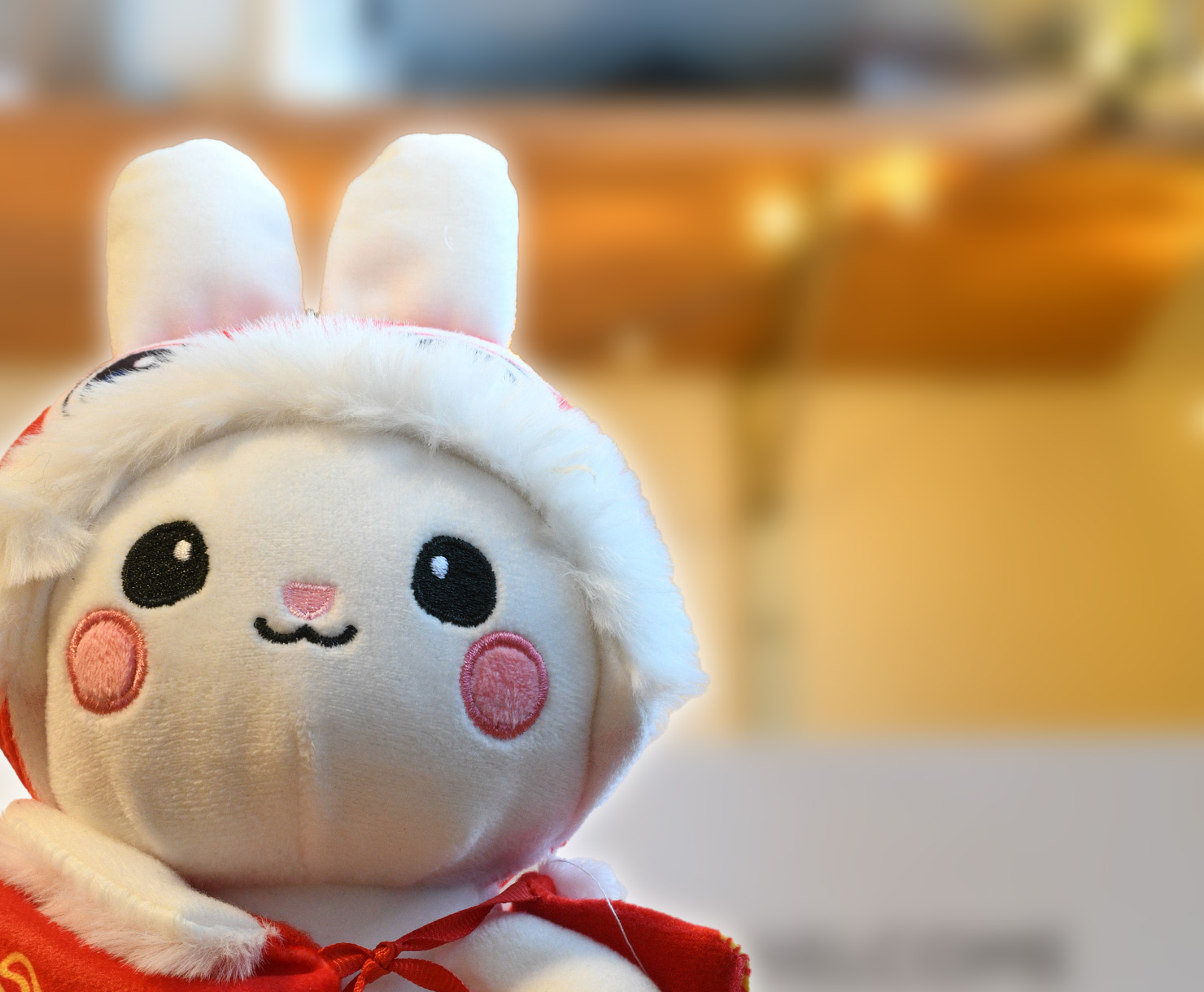}
    \caption{}
    \label{fig:intro-c}
  \end{subfigure}
  \caption{The effect of different aperture sizes on the bokeh effect and Gaussian blur. (a) With narrow aperture (f/16). (b) With wide aperture (f/1.8). (c) Artificial Gaussian blur.}
  \label{fig:intro}
\end{figure}

The bokeh effect refers to the aesthetic quality of the out-of-focus areas in a photograph. This effect is achieved by using a shallow depth of field in photography, which creates a blurred background while keeping the subject in focus. The bokeh effect requires a wide aperture lens, which is typically only available on high-end cameras with sophisticated optics and usually large sensors. Fig.~\ref{fig:intro-a} and Fig.~\ref{fig:intro-b} illustrate the impact of aperture size on the bokeh effect, whereby Fig.~\ref{fig:intro-a} was captured using a narrow aperture (f/16) and Fig.~\ref{fig:intro-b} was taken with a wide aperture (f/1.8). As these images show, a wide aperture produces a shallow depth-of-field with a blurry background, while a narrow aperture creates a deeper depth-of-field with a clearer background. This comparison underscores the significance of the aperture size in achieving desired bokeh effects. The bokeh effect is not identical to the blurring effect. For example, Fig.~\ref{fig:intro-c} demonstrates an artificial blurred image using Gaussian blur with the same blurring strength as Fig.~\ref{fig:intro-b}. However, the bokeh effect still can be simulated computationally using algorithms, making it possible to produce this effect on devices such as portable cameras or mobile phones without accessories. 

To learn the bokeh style directly from high-quality DSLR camera photos, a large-scale Everything is Better with Bokeh! (EBB!) dataset was proposed in \cite{ignatov2020rendering}, where the authors also built a deep learning-based method (namely PyNET) capable of transforming wide to shallow depth-of-field images automatically. However, this method relies on an accurate depth map estimation of an image, which is time-consuming and computationally costly. To address it, several works \cite{hariharanbokeh,dutta2021stacked,zheng2022constrained} propose to render bokeh effects in an end-to-end manner without relying on depth maps. By leveraging fancy network structures (\textit{e.g.}, Hierarchical Network \cite{dutta2021stacked} and Transformer \cite{hariharanbokeh}), they can achieve fast and realistic results on the EBB! dataset. However, their approaches have not yet gained widespread adoption due to the dataset limitation, where the bokeh effect transformation is fixed (on both lens models and aperture sizes) and thus cannot fit the requirements of real-world applications.

More specifically, the EBB! dataset has collected 5,000 pairs of aligned images captured by a Canon 70D DSLR camera. For each training pair, the source image is captured using a narrow aperture (f/16) and the target is captured using a wide aperture (f/1.8). This means the only bokeh transformation is to render blurry backgrounds from sharp photos. Based on this dataset, many solutions were presented to address the bokeh effect rendering problem \cite{ignatov2020aim,dutta2021stacked,peng2022bokehme}.

However, as we mentioned above, a major limitation of this dataset is the fixed lens and aperture parameters, meaning that there is little variation in the types of scenes and objects captured. 

To address it, NTIRE 2023 Bokeh Effect Transformation challenge~\cite{conde2023ntire_bokeh} proposes a new dataset, Bokeh Effect Transformation Dataset (BETD)~\cite{conde2023ntire_bokeh}, which contains image pairs with an artificially synthesized foreground portrait, a background captured by different lenses (Sony50mm and Canon50mm) and aperture sizes. Moreover, this dataset also provides foreground masks and metadata which contains lens parameters and the extra disparity values measuring how much two images of the same scene differ. One potential advantage of this dataset is that it provides a wider range of blur strengths than the EBB! Dataset, which is essential for real-world scenarios. Additionally, the inclusion of foreground masks can also be helpful for advanced training losses or evaluations.

Our work is built upon BETD, with full consideration of the metadata and foreground masks in a comprehensive manner. Specifically, we propose to embed the metadata in image inputs and assemble them into the network using affine transformations. Further, we leverage the alpha masks provided in the dataset and propose a novel loss-calculating method.

Our primary contributions are summarized as follows:

•	We propose a novel model, named \textit{BokehOrNot}, based on Restormer~\cite{zamir2022restormer}. Our model is capable of transforming and rendering the natural bokeh effect among multiple aperture sizes and lenses without compromising the sharpness of the foreground and the global image quality.

•	We build a Lens Embedding Module (LEM) and a Dual-Input Transformer Block (DITB) which integrate lens metadata to original image inputs with sinusoidal embedding and affine transforms, allowing the model to learn different bokeh styles and strengths efficiently. 

•	Alpha-masked loss that calculates loss excluding foreground areas. This helps the model to learn bokeh effect transformation more accurately without the influence of the foreground, especially with the cropping training technique.

\section{Related Work}
\label{sec:related_work}

\subsection{Bokeh Rendering}

Bokeh rendering is growing to be an important technique in computational photography, which usually takes photos with out-of-focus areas to highlight regions of interest. Hach \textit{et al}.~\cite{hach2015cinematic} introduced a Point-Spread-Function based approach to render different styles of bokeh. However, this method requires high-quality lens point-spread-function measurements and specific cameras. To render bokeh effect with minimal requirements, Ignatov \textit{et al}.~\cite{ignatov2020rendering} proposed a large-scale \textit{Everything is Better with Bokeh!} (EBB!) dataset which contains 5,000 aligned shallow / wide depth-of-field image pairs collected in the wild with the Canon 70D DSLR camera and 50mm f/1.8 fast lens. Subsequently, abundant works were proposed to tackle the bokeh rendering problem based on the EBB! dataset~\cite{qian2020bggan,dutta2021stacked,hariharanbokeh,zheng2022constrained,ignatov2023realistic}. Specifically, PyNET~\cite{ignatov2020rendering} is the first method proposed along with the EBB! dataset, which utilizes a multi-scale CNN architecture and meanwhile adds the pre-computed depth map to the input. Luo \textit{et al}.~\cite{luo2020bokeh} proposed a method based on defocus map estimation with radiance and upsampling to learn shallow depth-of-ﬁeld from a single bokeh-free image. BGGAN~\cite{qian2020bggan} combines a Glass-Net with Generative Adversarial Network (GAN) to improve the visual effect of synthetic bokeh rendering. To get rid of the requirement of depth estimation, Dutta \textit{et al}.~\cite{dutta2021stacked} created a Stacked DMSHN which does not depend on the depth map estimation and incorporates multiple encoder-decoder structures to gradually refine the output, resulting in an efficient bokeh rendering model. Moreover, Hariharan \textit{et al}.~\cite{hariharanbokeh} also proposed a depth map free transformer-based approach that is built on the PyNET but achieves better results in retaining the details of the foreground, namely BRViT. Moreover, BokehMe~\cite{peng2022bokehme} and AMPN~\cite{georgiadis2023adaptive} were proposed to generate photo-realistic bokeh effects. In particular, BokehMe utilizes imperfect disparity maps and can adjust blur sizes, focal planes, and aperture shapes. AMPN employs a mask prediction module which enables users to control their bokeh rendering with a user-guided mask.

\subsection{Vision Transformer and Image Restoration Network}

Vision transformer (ViT) has shown surprising performance in both high-level computer vision tasks~\cite{dosovitskiy2020image,liu2021swin,bao2021beit,strudel2021segmenter} and low-level image restoration tasks~\cite{liang2021swinir,zamir2022restormer,wang2022uformer,luo2022bsrt,chen2021pre}. IPT~\cite{chen2021pre} is the first method that employs ViT in image restoration, but it is not practical since the complexity of performing full self-attention is quadratic to the image size. To meet the requirement of limited GPU resources and a fast inference, SwinIR~\cite{liang2021swinir} successfully combines Swin Transformer~\cite{liu2021swin} with residual convolution layers to perform self-attention on large feature maps and achieves a reliable performance on several image restoration tasks. Meanwhile, Uformer~\cite{wang2022uformer} and Restormer~\cite{zamir2022restormer} were designed to address the problems that were usually solved by U-Net~\cite{ronneberger2015u} structures, such as denoising and deblurring. The former also applies local window attention and utilizes U-structure to capture global dependencies. And the latter applies self-attention across channels rather than the spatial dimension to make the complexity increase linearly. Meanwhile, lightweight networks like LPIENet~\cite{conde2023perceptual} remarkably reduce computational workload than state-of-the-art networks on image restoration tasks while maintaining close image quality, making it possible to run certain tasks on mobile devices.

\section{Dataset}
\label{sec:Dataset}
\begin{figure}
  \centering
  \begin{subfigure}{0.23\textwidth}
    \includegraphics[width=1.5in]{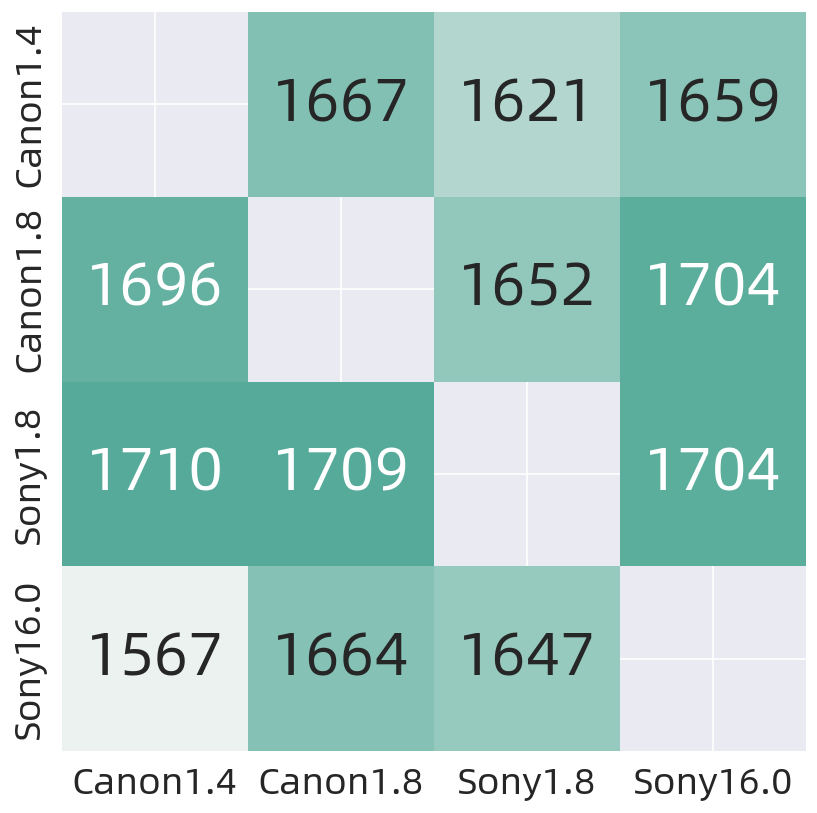}
    \caption{}
    \label{fig:dataset-a}
  \end{subfigure}
  \hfill
  \begin{subfigure}{0.23\textwidth}
    \includegraphics[width=1.5in]{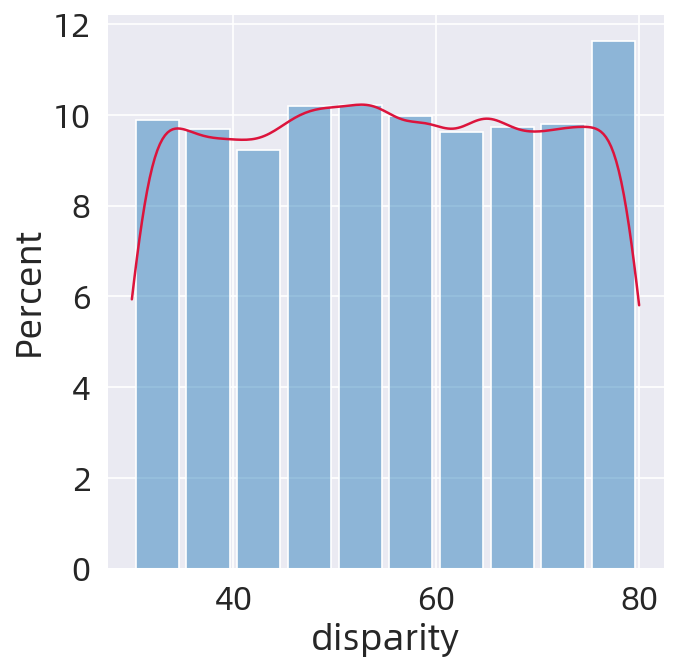}
    \caption{}
    \label{fig:dataset-b}
  \end{subfigure}
  \caption{(a): The transformation distribution of lens pairs in BETD~\cite{conde2023ntire_bokeh}. Each number in the grids indicates the number of transformation occurrences from the vertical lens labels to the horizontal lens labels. ``Canon1.4" is the abbreviation for ``Canon50mmf1.4BS", and so forth. (b): The percentages of different disparity values.}
  \label{fig:dataset}
\end{figure}

We assess our method on Bokeh Effect Transformation Dataset (BETD)~\cite{conde2023ntire_bokeh}, which consists of 20,000 training image sets and 500 validation images. Each image has a size of $1920 \times 1440$ pixels and each training pair includes a source image, a target image, and a metadata tuple of id, source lens, target lens and disparity value. The source and target images have the same foreground and background, but the latter is taken under a different set of lenses and apertures. Moreover, each training pair is accompanied by an alpha mask, which can be utilized for more advanced training losses or evaluations. Disparity value indicates the level of blur strength difference between the target image and the source image. To explore the dataset in more detail, we examined the dataset features and statistics. We create cross-tabulations (Fig.~\ref{fig:dataset-a}) for each pair of transformation combinations and plot the distribution of disparity (Fig.~\ref{fig:dataset-b}). Furthermore, we summarized the evaluation result by using the three following evaluation metrics, which are also used in our experiments to evaluate the performance of models:

\begin{itemize}
    \item[(i)]
    Peak Signal-to-Noise Ratio (PSNR). By calculating the ratio of the maximum possible power of a pixel to the power of corrupting noise. A higher PSNR indicates better image quality.
    \item[(ii)]
    Structual Similarity (SSIM)~\cite{wang2004image}. This metric quantifies the similarity between two images. A higher SSIM ensures the image is similar to its reference. SSIM is significant especially for blur-to-sharp bokeh transformation.
    \item[(iii)]
    Learned Perceptual Image Patch Similarity (LPIPS)~\cite{zhang2018unreasonable}. Unlike metrics such as PSNR or SSIM, LPIPS considers the human perception of image quality, making the evaluation system more comprehensive.
\end{itemize}

\begin{table}[htbp]
  \centering
  \begin{tabular}{lccc}
    \toprule
    Dataset & PSNR & SSIM & LPIPS \\
    \midrule
    BETD & 34.691 & 0.903 & 0.162 \\
    \bottomrule
  \end{tabular}
  \caption{Overall statistics of BETD~\cite{conde2023ntire_bokeh}.}
  \label{tab:BETD}
\end{table}

\begin{table}[htbp]
  \centering
  \resizebox{\linewidth}{!}{
  \begin{tabular}{l|cccc}  
    \toprule
    \diagbox[width=5.1em]{Source}{Target} & Canon1.4 & Canon1.8 & Sony1.8 & Sony16 \\
    \hline
    Canon1.4 & - & 46.013 & 42.062 & 24.293\\
    Canon1.8 & 46.016 & - & 45.624 & 24.623\\
    Sony1.8 & 42.030 & 45.693 & - & 25.052\\
    Sony16 & 24.431 & 24.741 & 24.995 & -\\
    \Xhline{1pt}
  \end{tabular}}
  \caption{PSNR of different transformation pairs}
  \label{tab:PSNR}
\end{table}

\begin{table}[htbp]
  \centering
  \begin{tabular}{l|cccc}
    \toprule
    \diagbox[width=5.1em]{Source}{Target} & Canon1.4 & Canon1.8 & Sony1.8 & Sony16 \\
    \hline
    Canon1.4 & - & 0.993 & 0.989 & 0.808\\
    Canon1.8 & 0.993 & - & 0.992 & 0.810\\
    Sony1.8 & 0.989 & 0.993 & - & 0.821\\
    Sony16 & 0.809 & 0.814 & 0.817 & -\\
    \Xhline{1pt}
  \end{tabular}
  \caption{SSIM~\cite{wang2004image} of different transformation pairs}
  \label{tab:SSIM}
\end{table}

\begin{table}[htbp]
  \centering
  \begin{tabular}{l|cccc}
    \toprule
    \diagbox[width=5.1em]{Source}{Target} & Canon1.4 & Canon1.8 & Sony1.8 & Sony16 \\
    \hline
    Canon1.4 & - & 0.005 & 0.011 & 0.324\\
    Canon1.8 & 0.005 & - & 0.007 & 0.319\\
    Sony1.8 & 0.011 & 0.007 & - & 0.311\\
    Sony16 & 0.327 & 0.318 & 0.315 & -\\
    \Xhline{1pt}
  \end{tabular}
  \caption{LPIPS~\cite{zhang2018unreasonable} of different transformation pairs}
  \label{tab:LPIPS}
\end{table}

Table~\ref{tab:BETD} shows the overall statistics of BETD~\cite{conde2023ntire_bokeh}. All values are calculated between source images and the corresponding target images. Table~\ref{tab:PSNR},\ref{tab:SSIM} and \ref{tab:LPIPS} illustrate the statistics on different transformations. These statistics can be referenced as baselines to evaluate model performance.

\section{Proposed Framework}
\label{sec:framework}
\begin{figure*}[ht] 
  \centering
  \includegraphics[width=1.\linewidth]{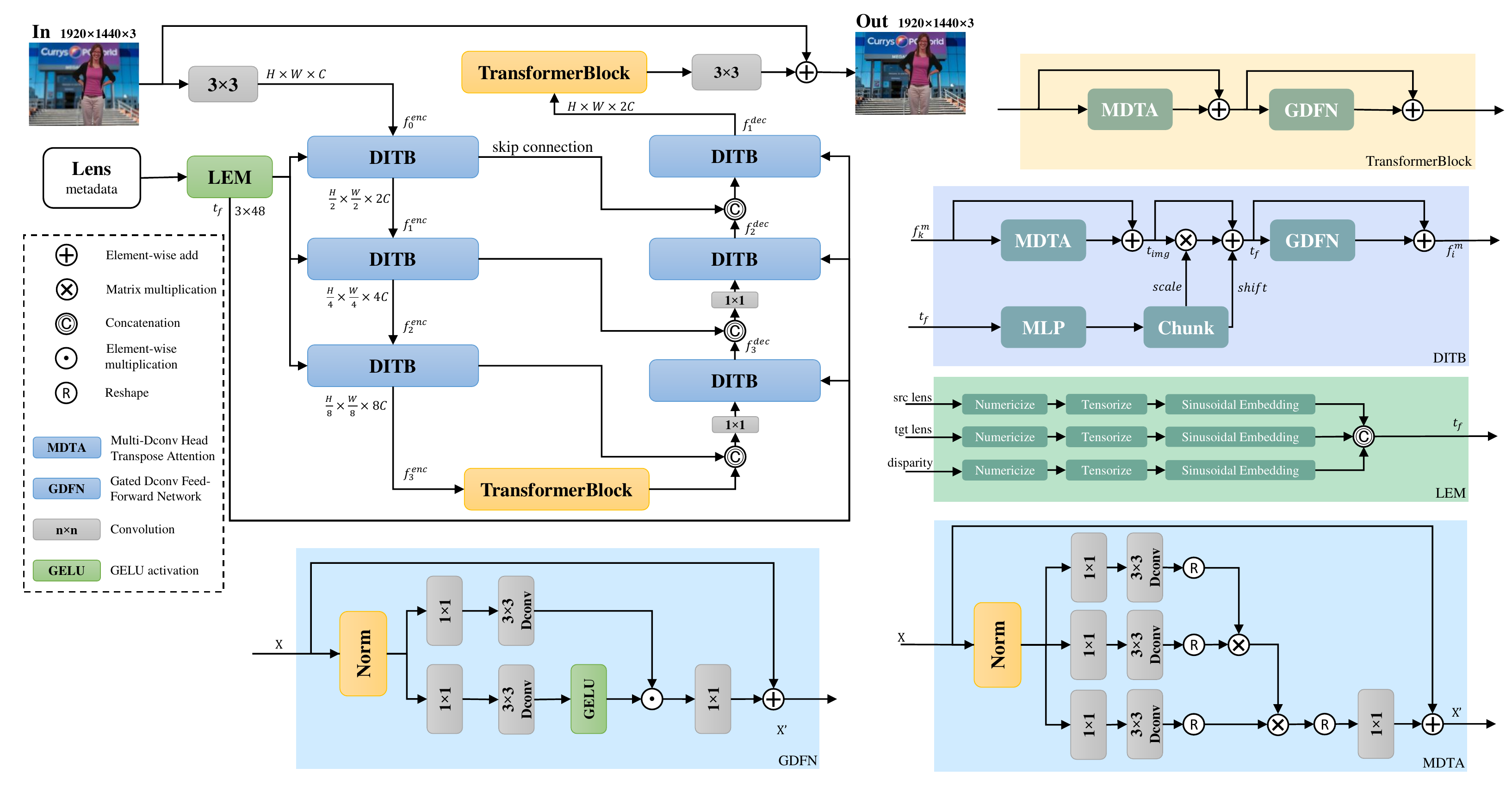}
  \caption{Overview of our BokehOrNot Framework. Our BokehOrNot consists of multi-scale U-net~\cite{ronneberger2015u} design incorporating dual-input transformer block (DITB) and lens embedding module (LEM). GDFN and MDTA are blocks adopted from~\cite{zamir2022restormer}.}
  \label{fig:vec_framework}
\end{figure*}
The bokeh effect, which involves sophisticated factors, is an optical phenomenon generated by the light passing through the optical module of the lens. Bokeh effects gain explicit differences from multiple aspects, including the sharpness of the bokeh, the aberration around the bokeh, the distortion of the bokeh \textit{etc.} 

\textbf{Overview.} The proposed BokehOrNot framework is based on the Restormer~\cite{zamir2022restormer} framework, a Transformer-based network, and combined with factors related to bokeh effect. BokehOrNot contains a lens information embedding module and a symmetric encoder-decoder structure with dual-input Transformer blocks. Our proposed model framework is shown in Fig.~\ref{fig:vec_framework}.

\subsection{Lens Embedding Module (LEM)}
Optics and photography professionals have verified that different bokeh effects are caused by multiple factors within lenses\cite{sivokon2014theory}, including aperture sizes, the material of lens elements, the number of diaphragm blades \cite{nasse2010depth} \textit{etc.} Among these factors, the aperture size has been identified as a crucial factor. This observation has led to an approach in which aperture sizes and lens models are extracted from the BETD~\cite{conde2023ntire_bokeh} to leverage lens information.

To utilize the lens information, the aperture size, which significantly affects the intensity of the bokeh effect, is expressed as the value of the corresponding size, \textit{e.g.} a f/1.8 aperture will be expressed as 1.8. Furthermore, to distinguish lenses, we use one-hot encoding on the lens metadata by the following pattern:
\begin{equation} 
\label{eq:1}
meta = [{k_1},{k_2}, \cdots ,{k_{n-1}},aperture],
\end{equation}
where $n$ is the kinds of lenses, ${k_i} \in \{  - 1,1\}$ and ${k_i =1}$ for the $i$-th lens whereby the others are $-1$. Considering the new BETD~\cite{conde2023ntire_bokeh}, Sony lenses are assigned $[1,aperture]$ and Canon lenses are assigned $[-1,aperture]$. For example, a Canon f/1.4 lens is transformed into a value of $[-1,1.4]$ (under a binary scenario, -1.4 is identical to $[-1,1.4]$), allowing the model to distinguish lens models that significantly influence the bokeh style through an explicit value. By calculating the Hamming distance between the source lens tensor and the target lens tensor while a longer distance typically represents bigger differences, this encoding ensures that:
\begin{itemize}
    \item[(i)]
    When the source lens model and the target lens model are the same, the magnitude of the bokeh transformation is expressed by the absolute value of the difference between aperture values. A larger absolute value signifies a greater transformation magnitude. For example, a transformation from Sony50mm f/1.4 to Sony50mm f/1.8 has a difference of 0.4 while a transformation from Sony50mm f/1.8 to Sony50mm f/16.0 has a difference of 14.2. The latter applies to a larger transformation magnitude.
    \item[(ii)]
    When aperture sizes are same, the transformation gains the same magnitude between different lenses.
    \item[(iii)]
    Transformations involving both lens type and aperture size obtain a greater Hamming distance than changing a single factor alone.
\end{itemize}

The disparity, which is an indicator of blur strength, receives identical processing as the aperture size. This approach facilitates the indication of differences in bokeh intensity at various aperture sizes through the difference in values.

To learn the transformation from the source lens to the target lens with a certain disparity, we use sinusoidal positional embedding~\cite{vaswani2017attention} to process source lens values, target lens values and disparity values. Each of these values is ``tensorized" to an $1 \times 48$  float tensor and three tensors are then concatenated into a $3 \times 48$ tensor. A two-layer perceptron helps to extract features from this tensor and returns an $1 \times 48$ tensor.

\subsection{Dual-Input Transformer Block (DITB)}
Upon obtaining the embedding tensor from the LEM, we incorporate lens information into the network forwarding process. To this end, a dual-input transformer block based on vision transformer~\cite{dosovitskiy2020image} is introduced to leverage both the image input and the lens input. The lens input is subject to processing by a two-layer perceptron, which produces an intermediate tensor $t_p$ with dimensions $2 \cdot dim$. Subsequently, $t_p$ is chunked into two distinct segments, namely the $scale$ and $shift$. The image input $t_{img}$ is only updated by multi-Dconv head transposed attention (MDTA)~\cite{zamir2022restormer} in this phase.

\begin{equation} 
\label{eq:2}
t_{img} = t_{img} + MDTA(t_{img})
\end{equation}

In order to achieve a comprehensive utilization of lens information, we propose a novel approach that involves the fusion of both lens and image inputs, enhanced by scaling and shifting operations. The outcome of this process $t_f$ is a fused feature that encapsulates a more holistic representation of the input data:

\begin{equation} \label{eq:3}
t_f = t_{img} \cdot (1+scale) + shift
\end{equation}
which makes key information propagate further, gated-Dconv feed-forward network (GDFN)~\cite{zamir2022restormer} is applied to the $t_f$, with residual added after. The output of DITB is a feature map with a size corresponding to the depth of the block:

\begin{equation} \label{eq:4}
f_i^{m} = t_f + GDFN(t_f)
\end{equation}
where $i \in \{1,2,3\}$ and $m \in \{enc, dec\}$.

\subsection{Alpha-masked Loss}
Our proposed framework employs $L_1$ loss function and $L_1$ loss with foreground area removed in different training stages. Specifically, when performing bokeh effect transformation, the foreground, \textit{e.g.} people, pets, and other salient objects, should not be altered to achieve a higher verisimilitude. Along with the BETD~\cite{conde2023ntire_bokeh}, it motivates us to exclude the foreground area which is covered by the alpha mask when calculating loss. 

\begin{figure}
  \centering
  \includegraphics[scale=0.3]{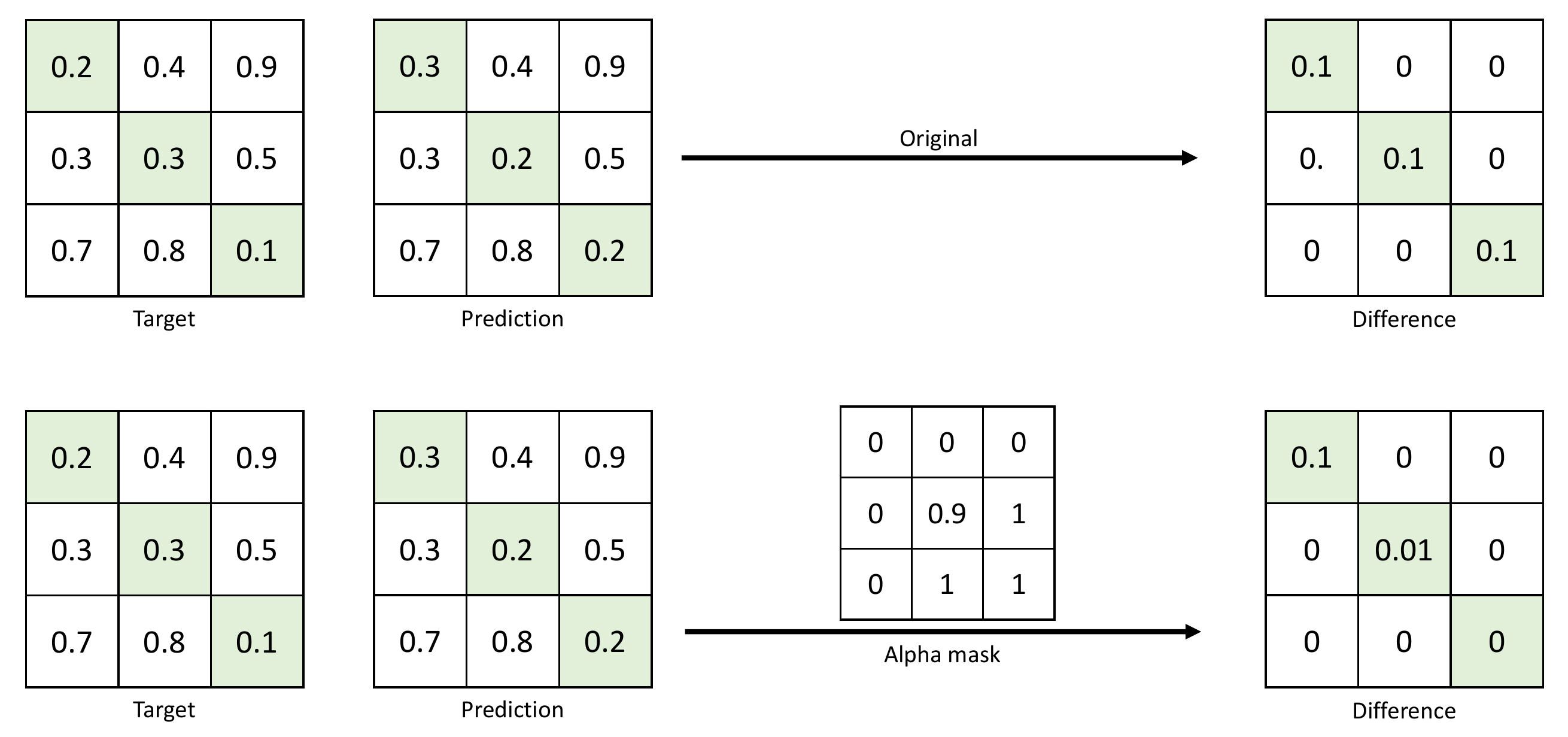}
  \caption{Demonstration of the alpha-masked loss. The loss of background area (upper-left grid in this example) obtains a larger proportion of the loss. The edge area (central grid) still contributes to the loss with less significance while the foreground area (lower-right grid) is no longer considered.}
  \label{fig:loss}
\end{figure}

Figure~\ref{fig:loss} illustrates how the alpha-masked loss is generated. Grids in green represent pixels that failed to predict accurately. In the original $L_1$ loss, only one of the three green pixels contributes to transformation in the background and all areas contribute equally, while in alpha-masked loss, green grids in the foreground area are no longer considered. The loss now involves only the background area and slightly considers the edge area.

The alpha-masked loss can be represented as:
\begin{equation} \label{eq:5}
L = \frac{1}{N}\sum\limits_{i = 1}^N {\left| {pre{d_i} - g{t_i}} \right|}  * (1 - alpha_i)
\end{equation}
where $pred_i$ is the $i$-th pixel of the transformation result from our model, $alpha$ is the mask from BETD~\cite{conde2023ntire_bokeh}, $gt$ is the ground truth and $N$ is the number of pixels. The alpha-masked loss performs element-wise multiplication between the alpha mask and the prediction image or the target image, resulting in a zero-filled area. 

With this, the learning process benefits from two perspectives: alleviating the transformation on the foreground area and eliminating the impact of loss increment due to the foreground area while learning bokeh effect within background areas.

\section{Experimental Results}

\subsection{Implementation Details}
BokehOrNot is implemented using PyTorch and trained with the Adam~\cite{kingma2014adam} optimizer. Two training stages are involved: the precise detecting stage and the global transformation stage.

During the precise detecting stage, the batch size is set to 4 with a learning rate of 1e-4. The loss function is the original $L_1$ loss, with an input image size randomly cropped to $256 \times 256$. The global transformation stage takes advantage of our alpha-masked loss shown in Eq.~\ref{eq:5} with $384 \times 384$ input size, enabling the model to learn holistically and ignoring the foreground. The learning rate is set to 5e-5 and the batch size is reduced to 2. The total training time is 70 hours, with approximately 35 hours for each stage. The training is performed on 2 NVIDIA A100 40GB GPUs. An average inference time for a $1920 \times 1440$ size input is 0.94s on a single NVIDIA A100 40GB GPU.

\begin{figure*}
  \centering
  \includegraphics[scale=0.5]{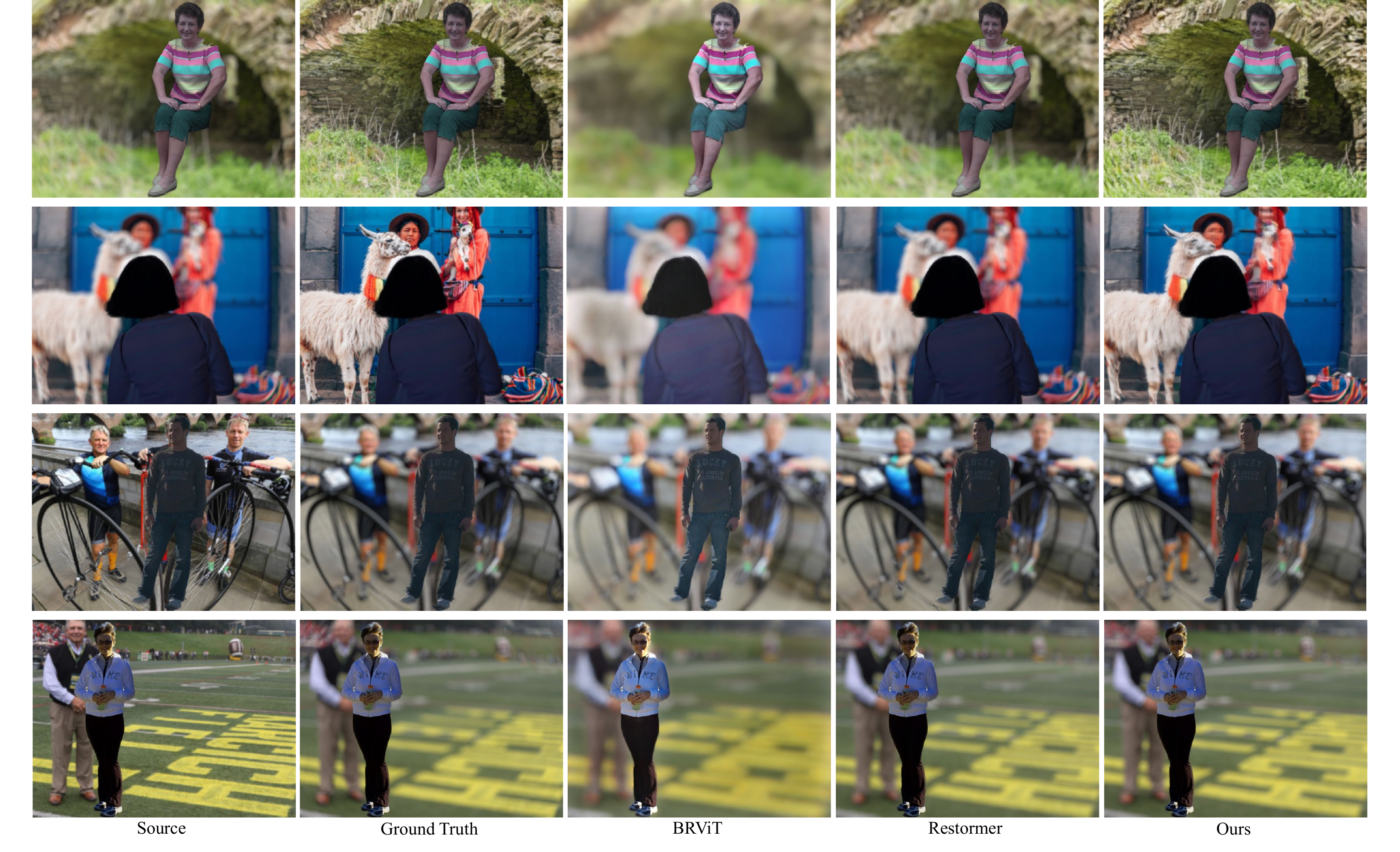}
  \caption{Qualitative comparison of bokeh effect transformation. The first image applies a transformation from Sony50mmf1.8BS to Sony50mmf16.0BS and the second image applies a transformation from Canon50mmf1.4BS to Sony50mmf16.0BS, \textit{i.e.} blur-to-sharp. The following two images are transformed from Sony50mmf16.0BS to Sony50mmf1.8BS and Canon50mmf1.4BS, respectively. BRViT~\cite{hariharanbokeh} failed to render realistic bokeh effect and Restormer~\cite{zamir2022restormer} cannot render blur-to-sharp transformation ideally.}
  \label{fig:compare_small}
\end{figure*}

\begin{figure*}
  \centering
  \includegraphics[scale=0.47]{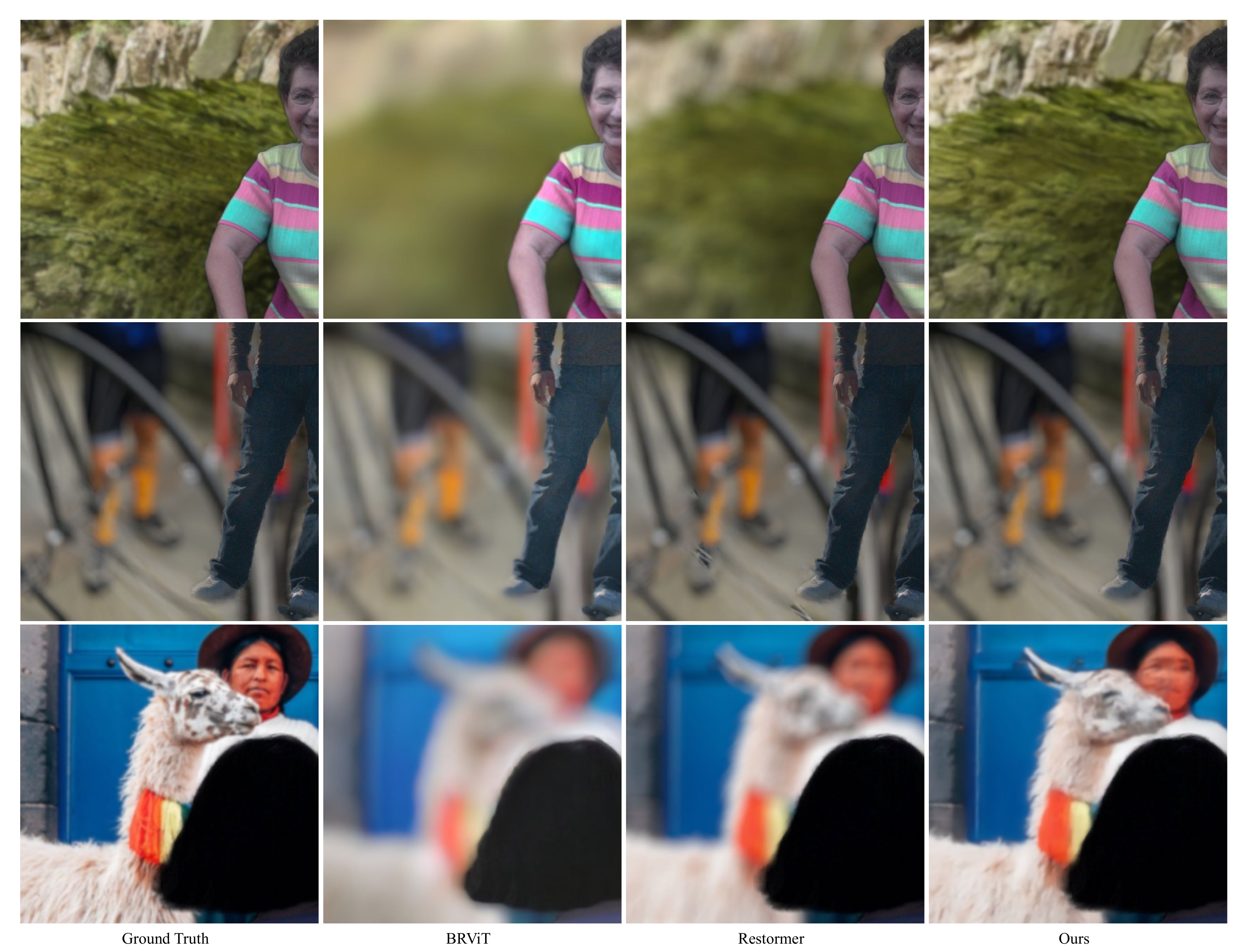}
  \caption{Magnified samples of typical transformations. BokehOrNot generates higher quality images than BRViT~\cite{hariharanbokeh} and Restormer~\cite{zamir2022restormer}.}
  \label{fig:compare_large}
\end{figure*}

\begin{table}
  \centering
  \begin{tabular}{lccc}
    \toprule
    Model & PSNR$\uparrow$ & SSIM$\uparrow$ & LPIPS$\downarrow$ \\
    \midrule
    BRViT(i)~\cite{hariharanbokeh} & 12.620 & 0.765 & 0.214\\
    Restormer~\cite{zamir2022restormer}(i) & 36.594 & 0.942 & 0.106\\
    Ours(i) & 38.090 &  0.984 & 0.041\\
    Ours(ii) & 39.320 &  0.967 & 0.049\\
    \bottomrule
  \end{tabular}
  \caption{Quantitative results on different models. (i) a total of 500 “f/16.0 to f/1.8” images are tested in the experiments, (ii) a total of 500 images of any bokeh transformation are tested.}
  \label{tab:models}
\end{table}

\begin{table}
  \centering
  \begin{tabular}{lccc}
    \toprule
    Model & PSNR$\uparrow$ & SSIM$\uparrow$ & LPIPS$\downarrow$ \\
    \midrule
    Restormer~\cite{zamir2022restormer} & 36.594 & 0.942 & 0.106\\
    Ours(i) & 38.534 &  0.965 & 0.058\\
    Ours(ii) & 38.558 & 0.966 & 0.051 \\
    Ours(iii) & 39.137 &  0.964 & 0.055\\
    Ours(iv)  & 39.320 &  0.967 & 0.049\\
    \bottomrule
  \end{tabular}
  \caption{Ablation experiments results on proposed approaches. The Restormer~\cite{zamir2022restormer} model applies the same settings in Table~\ref{tab:models}. (i) only with LEM+DITB, (ii) model (i) with alpha-masked loss, (iii) model (i) with enlarged inputs applied, (iv) proposed BokehOrNot model. A total of 500 images of any bokeh transformation are tested.}
  \label{tab:ablation}
\end{table}

\begin{table*}
\begin{center}
\begin{small}
\resizebox{.65\linewidth}{!}{
\begin{tabular}{lccccc}
\toprule
\multirow{2}{*}{Team} &  \multicolumn{3}{c}{All images} & \multicolumn{2}{c}{Real images}  \\ \cmidrule(lr){2-4} \cmidrule(lr){5-6}
&  PSNR$\uparrow$ & SSIM$\uparrow$ & LPIPS$\downarrow$ & SSIM$\uparrow$ & LPIPS$\downarrow$ \\
\midrule
Ours (BokehOrNot)  & 32.288 & 0.9327 & 0.1130 & 0.8423  & 0.2199\\
EBokehNet~\cite{seizinger2023bokeh} & 34.543 & 0.9350 & 0.1039 & 0.8414 & 0.2206\\
IR-SDE~\cite{luo2023refusion}  & 30.866 & 0.9297 & 0.1301 & 0.8427  & 0.2387\\
BIGbaodan  & 30.327 & 0.9281 & 0.1249 & 0.8415  & 0.2178\\
AIA-Smart  & 34.572 & 0.9361 & 0.0966 & 0.8435  & 0.2224\\
Samsung Research China  & 35.264 & 0.9362 & 0.0985 & 0.8416  & 0.2186\\
NUS-LV-Bokeh  & 32.326 & 0.9333 & 0.1076 & 0.8420  & 0.2230\\
IPAL-Bokeh  & 32.076 & 0.9324 & 0.1076 & 0.8419  & 0.2161\\
Baseline  & 28.599 & 0.9128 & 0.1878 & 0.8412  & 0.2181\\

\bottomrule
\end{tabular}
}
\end{small}
\end{center}
\vskip -0.1in
\caption{Quantitative results from all teams above baseline. This result is calculated on an undisclosed test set with both artificial synthetic images and real-captured images. A total of 155 images are tested, 95 are synthetic images and 60 are real-captured.}
\label{table:challenge}
\end{table*}

\subsection{Results and Analysis}
To the best of our knowledge, no publicly available deep network models now involve lens metadata. Therefore, we select two recently published network structures for bokeh effect rendering and image restoration with publicly available code as references. Considering current models are primarily focusing on rendering shallow depth-of-field from a rather sharp image, we preclude blur-to-sharp transformations in the dataset. All models are trained with recommended settings or similar settings as ours.

Our proposed BokehOrNot model outperforms on all three metrics. The LEM and DITB which utilize lens metadata contribute greatly to the evaluation results shown in Table~\ref{tab:models}. As bokeh effect transformation quality is the subjective perception of people to some extent, we also conduct qualitative comparisons, shown in Fig.~\ref{fig:compare_small}. 

BRViT~\cite{hariharanbokeh} model blurs images under any scenario and engages brightness distortion. Restormer~\cite{zamir2022restormer} can distinguish the direction of transformation to some extent, \textit{i.e.} “to blur” or “to sharp”. However, it still cannot perform blur-to-sharp transformation with satisfactory quality. Our model renders both types of transformation desirably with natural bokeh effect.

Figure~\ref{fig:compare_large} shows magnified images with details. Comparing to BRViT~\cite{hariharanbokeh} and Restormer~\cite{zamir2022restormer}, our model significantly restores the texture of rocks and mosses while distinguishing a clear edge between the background and the foreground in the first image. In the second image, the Restormer~\cite{zamir2022restormer} model failed to render the bokeh effect on all background areas, especially the area adjacent to wheel spokes, whereby our model renders natural bokeh effect. In general, our model is capable of rendering bokeh effects under different scenarios while retaining a sharp foreground, creating a natural and pleasing visual effect.

\subsection{Ablation Study}
As we integrate multiple new approaches in our framework, ablation experiments are conducted to further verify and evaluate the performance and robustness of these methods, including: applying LEM and DITB, applying alpha-masked loss and applying enlarged inputs. 

Table~\ref{tab:ablation} indicates our proposed methods still improve the performance when applied separately. LEM and DITB notably elevate all metrics in comparison with Restormer~\cite{zamir2022restormer}. The incorporation of alpha-masked loss contributes notably to a reduction in the LPIPS~\cite{zhang2018unreasonable} metric and the enlarged inputs method results in a significant enhancement in the PSNR metric.

\section{NTIRE 2023 Challenge}
We enroll in NTIRE 2023 Bokeh Transformation Challenge~\cite{conde2023ntire_bokeh}. The model proposed in this paper participates in the joint comparison among different models. Quantitative results are provided by challenge organizers, shown in Table~\ref{table:challenge}. The result contains two parts: all images including synthesized images, and only real images. Real images are taken directly from a camera, and only apply transformations between f/1.4 and f/1.8.

In this challenge, we use the direct output of the model, without any model ensemble the test-time augmentation. Our result surpasses the baseline for 3.689 dB on PSNR, ranking the fifth place in this competition.

\section{Conclusion}
Bokeh Effect Transformation Dataset~\cite{conde2023ntire_bokeh} is the first dataset ever incorporating lens metadata and multi-style bokeh effect transformation. We introduce the BokehOrNot model with two new methods to transform and render beautiful bokeh effect, namely lens embedding module with dual-input Transformer block and alpha-masked loss, along with this novel dataset. The former focuses on integrating lens metadata efficiently, allowing the model to learn different bokeh styles, while the latter aims to learn specific bokeh effect more accurately. Quantitative and qualitative experiments demonstrate that our proposed model can transform and render natural bokeh effect given metadata and outperforms cutting-edge models on three metrics. Considering images taken by either portable cameras or mobile phones are embedded with lens metadata, it is feasible to apply this model to real-world scenarios.

\section*{Acknowledgements}
The computations were enabled by the \textit{Berzelius} resources provided by the Knut and Alice Wallenberg Foundation at the National Supercomputer Centre. And thanks to Ziwei Luo for insightful feedback.

{\small
\bibliographystyle{ieee_fullname}
\bibliography{egbib}

\begin{thebibliography}{10}\itemsep=-1pt

\bibitem{bao2021beit}
Hangbo Bao, Li Dong, Songhao Piao, and Furu Wei.
\newblock Beit: Bert pre-training of image transformers.
\newblock {\em arXiv preprint arXiv:2106.08254}, 2021.

\bibitem{chen2021pre}
Hanting Chen, Yunhe Wang, Tianyu Guo, Chang Xu, Yiping Deng, Zhenhua Liu, Siwei
  Ma, Chunjing Xu, Chao Xu, and Wen Gao.
\newblock Pre-trained image processing transformer.
\newblock In {\em Proceedings of the IEEE/CVF Conference on Computer Vision and
  Pattern Recognition}, pages 12299--12310, 2021.

\bibitem{conde2023ntire_bokeh}
Marcos~V Conde, Manuel Kolmet, Tim Seizinger, Tom~E Bishop, and Radu Timofte.
\newblock Lens-to-lens bokeh effect transformation. ntire 2023 challenge
  report.
\newblock In {\em Proceedings of the IEEE/CVF Conference on Computer Vision and
  Pattern Recognition Workshops}, 2023.

\bibitem{conde2023perceptual}
Marcos~V Conde, Florin Vasluianu, Javier Vazquez-Corral, and Radu Timofte.
\newblock Perceptual image enhancement for smartphone real-time applications.
\newblock In {\em Proceedings of the IEEE/CVF Winter Conference on Applications
  of Computer Vision}, pages 1848--1858, 2023.

\bibitem{dosovitskiy2020image}
Alexey Dosovitskiy, Lucas Beyer, Alexander Kolesnikov, Dirk Weissenborn,
  Xiaohua Zhai, Thomas Unterthiner, Mostafa Dehghani, Matthias Minderer, Georg
  Heigold, Sylvain Gelly, et~al.
\newblock An image is worth 16x16 words: Transformers for image recognition at
  scale.
\newblock {\em arXiv preprint arXiv:2010.11929}, 2020.

\bibitem{dutta2021stacked}
Saikat Dutta, Sourya~Dipta Das, Nisarg~A Shah, and Anil~Kumar Tiwari.
\newblock Stacked deep multi-scale hierarchical network for fast bokeh effect
  rendering from a single image.
\newblock In {\em Proceedings of the IEEE/CVF Conference on Computer Vision and
  Pattern Recognition}, pages 2398--2407, 2021.

\bibitem{georgiadis2023adaptive}
Konstantinos Georgiadis, Albert Sa{\`a}-Garriga, Mehmet~Kerim Yucel, Anastasios
  Drosou, and Bruno Manganelli.
\newblock Adaptive mask-based pyramid network for realistic bokeh rendering.
\newblock In {\em Computer Vision--ECCV 2022 Workshops: Tel Aviv, Israel,
  October 23--27, 2022, Proceedings, Part II}, pages 429--444. Springer, 2023.

\bibitem{hach2015cinematic}
Thomas Hach, Johannes Steurer, Arvind Amruth, and Artur Pappenheim.
\newblock Cinematic bokeh rendering for real scenes.
\newblock In {\em Proceedings of the 12th European Conference on Visual Media
  Production}, pages 1--10, 2015.

\bibitem{ignatov2020rendering}
Andrey Ignatov, Jagruti Patel, and Radu Timofte.
\newblock Rendering natural camera bokeh effect with deep learning.
\newblock In {\em Proceedings of the IEEE/CVF Conference on Computer Vision and
  Pattern Recognition Workshops}, pages 418--419, 2020.

\bibitem{ignatov2020aim}
Andrey Ignatov, Radu Timofte, Ming Qian, Congyu Qiao, Jiamin Lin, Zhenyu Guo,
  Chenghua Li, Cong Leng, Jian Cheng, Juewen Peng, et~al.
\newblock Aim 2020 challenge on rendering realistic bokeh.
\newblock In {\em Computer Vision--ECCV 2020 Workshops: Glasgow, UK, August
  23--28, 2020, Proceedings, Part III 16}, pages 213--228. Springer, 2020.

\bibitem{ignatov2023realistic}
Andrey Ignatov, Radu Timofte, Jin Zhang, Feng Zhang, Gaocheng Yu, Zhe Ma,
  Hongbin Wang, Minsu Kwon, Haotian Qian, Wentao Tong, et~al.
\newblock Realistic bokeh effect rendering on mobile gpus, mobile ai \& aim
  2022 challenge: report.
\newblock In {\em Computer Vision--ECCV 2022 Workshops: Tel Aviv, Israel,
  October 23--27, 2022, Proceedings, Part III}, pages 153--173. Springer, 2023.

\bibitem{kingma2014adam}
Diederik~P Kingma and Jimmy Ba.
\newblock Adam: A method for stochastic optimization.
\newblock {\em arXiv preprint arXiv:1412.6980}, 2014.

\bibitem{liang2021swinir}
Jingyun Liang, Jiezhang Cao, Guolei Sun, Kai Zhang, Luc Van~Gool, and Radu
  Timofte.
\newblock Swinir: Image restoration using swin transformer.
\newblock In {\em Proceedings of the IEEE/CVF international conference on
  computer vision}, pages 1833--1844, 2021.

\bibitem{liu2021swin}
Ze Liu, Yutong Lin, Yue Cao, Han Hu, Yixuan Wei, Zheng Zhang, Stephen Lin, and
  Baining Guo.
\newblock Swin transformer: Hierarchical vision transformer using shifted
  windows.
\newblock In {\em Proceedings of the IEEE/CVF international conference on
  computer vision}, pages 10012--10022, 2021.

\bibitem{luo2020bokeh}
Xianrui Luo, Juewen Peng, Ke Xian, Zijin Wu, and Zhiguo Cao.
\newblock Bokeh rendering from defocus estimation.
\newblock In {\em Computer Vision--ECCV 2020 Workshops: Glasgow, UK, August
  23--28, 2020, Proceedings, Part III 16}, pages 245--261. Springer, 2020.

\bibitem{luo2023refusion}
Ziwei Luo, Fredrik~K Gustafsson, Zheng Zhao, Jens Sj{\"o}lund, and Thomas~B
  Sch{\"o}n.
\newblock Refusion: Enabling large-size realistic image restoration with
  latent-space diffusion models.
\newblock In {\em Proceedings of the IEEE/CVF Conference on Computer Vision and
  Pattern Recognition}, pages 1680--1691, 2023.

\bibitem{luo2022bsrt}
Ziwei Luo, Youwei Li, Shen Cheng, Lei Yu, Qi Wu, Zhihong Wen, Haoqiang Fan,
  Jian Sun, and Shuaicheng Liu.
\newblock Bsrt: Improving burst super-resolution with swin transformer and
  flow-guided deformable alignment.
\newblock In {\em Proceedings of the IEEE/CVF Conference on Computer Vision and
  Pattern Recognition}, pages 998--1008, 2022.

\bibitem{hariharanbokeh}
Hariharan Nagasubramaniam and Rabih Younes.
\newblock Bokeh effect rendering with vision transformers.
\newblock {\em TechRxiv preprint techrxiv.17714849.v1}, 2022.

\bibitem{nasse2010depth}
HH Nasse.
\newblock Depth of field and bokeh.
\newblock {\em Carl Zeiss camera lens division report}, 1, 2010.

\bibitem{peng2022bokehme}
Juewen Peng, Zhiguo Cao, Xianrui Luo, Hao Lu, Ke Xian, and Jianming Zhang.
\newblock Bokehme: When neural rendering meets classical rendering.
\newblock In {\em Proceedings of the IEEE/CVF Conference on Computer Vision and
  Pattern Recognition}, pages 16283--16292, 2022.

\bibitem{qian2020bggan}
Ming Qian, Congyu Qiao, Jiamin Lin, Zhenyu Guo, Chenghua Li, Cong Leng, and
  Jian Cheng.
\newblock Bggan: Bokeh-glass generative adversarial network for rendering
  realistic bokeh.
\newblock In {\em Computer Vision--ECCV 2020 Workshops: Glasgow, UK, August
  23--28, 2020, Proceedings, Part III 16}, pages 229--244. Springer, 2020.

\bibitem{ronneberger2015u}
Olaf Ronneberger, Philipp Fischer, and Thomas Brox.
\newblock U-net: Convolutional networks for biomedical image segmentation.
\newblock In {\em Medical Image Computing and Computer-Assisted
  Intervention--MICCAI 2015: 18th International Conference, Munich, Germany,
  October 5-9, 2015, Proceedings, Part III 18}, pages 234--241. Springer, 2015.

\bibitem{seizinger2023bokeh}
Tim Seizinger, Marcos~V Conde, Manuel Kolmet, Tom~E Bishop, and Radu Timofte.
\newblock Efficient multi-lens bokeh effect rendering and transformation.
\newblock In {\em Proceedings of the IEEE/CVF Conference on Computer Vision and
  Pattern Recognition Workshops}, 2023.

\bibitem{sivokon2014theory}
Viktor~P Sivokon and Michael~D Thorpe.
\newblock Theory of bokeh image structure in camera lenses with an aspheric
  surface.
\newblock {\em Optical Engineering}, 53(6):065103--065103, 2014.

\bibitem{strudel2021segmenter}
Robin Strudel, Ricardo Garcia, Ivan Laptev, and Cordelia Schmid.
\newblock Segmenter: Transformer for semantic segmentation.
\newblock In {\em Proceedings of the IEEE/CVF international conference on
  computer vision}, pages 7262--7272, 2021.

\bibitem{vaswani2017attention}
Ashish Vaswani, Noam Shazeer, Niki Parmar, Jakob Uszkoreit, Llion Jones,
  Aidan~N Gomez, {\L}ukasz Kaiser, and Illia Polosukhin.
\newblock Attention is all you need.
\newblock {\em Advances in neural information processing systems}, 30, 2017.

\bibitem{wang2004image}
Zhou Wang, Alan~C Bovik, Hamid~R Sheikh, and Eero~P Simoncelli.
\newblock Image quality assessment: from error visibility to structural
  similarity.
\newblock {\em IEEE transactions on image processing}, 13(4):600--612, 2004.

\bibitem{wang2022uformer}
Zhendong Wang, Xiaodong Cun, Jianmin Bao, Wengang Zhou, Jianzhuang Liu, and
  Houqiang Li.
\newblock Uformer: A general u-shaped transformer for image restoration.
\newblock In {\em Proceedings of the IEEE/CVF Conference on Computer Vision and
  Pattern Recognition}, pages 17683--17693, 2022.

\bibitem{zamir2022restormer}
Syed~Waqas Zamir, Aditya Arora, Salman Khan, Munawar Hayat, Fahad~Shahbaz Khan,
  and Ming-Hsuan Yang.
\newblock Restormer: Efficient transformer for high-resolution image
  restoration.
\newblock In {\em Proceedings of the IEEE/CVF Conference on Computer Vision and
  Pattern Recognition}, pages 5728--5739, 2022.

\bibitem{zhang2018unreasonable}
Richard Zhang, Phillip Isola, Alexei~A Efros, Eli Shechtman, and Oliver Wang.
\newblock The unreasonable effectiveness of deep features as a perceptual
  metric.
\newblock In {\em Proceedings of the IEEE conference on computer vision and
  pattern recognition}, pages 586--595, 2018.

\bibitem{zheng2022constrained}
Bolun Zheng, Quan Chen, Shanxin Yuan, Xiaofei Zhou, Hua Zhang, Jiyong Zhang,
  Chenggang Yan, and Gregory Slabaugh.
\newblock Constrained predictive filters for single image bokeh rendering.
\newblock {\em IEEE Transactions on Computational Imaging}, 8:346--357, 2022.

\end{thebibliography}
}

\end{document}